# Analysis of Diffractive Optical Neural Networks and Their Integration with Electronic Neural Networks

Deniz Mengu, Yi Luo, Yair Rivenson, and Aydogan Ozcan

*Abstract*—Optical machine learning offers advantages in terms of power efficiency, scalability and computation speed. Recently, an optical machine learning method based on Diffractive Deep Neural Networks ($D^2$NNs) has been introduced to execute a function as the input light diffracts through passive layers, designed by deep learning using a computer. Here we introduce improvements to $D^2$NNs by changing the training loss function and reducing the impact of vanishing gradients in the error back-propagation step. Using five phase-only diffractive layers, we numerically achieved a classification accuracy of 97.18% and 89.13% for optical recognition of handwritten digits and fashion products, respectively; using both phase and amplitude modulation (complex-valued) at each layer, our inference performance improved to 97.81% and 89.32%, respectively. Furthermore, we report the integration of $D^2$NNs with electronic neural networks to create hybrid-classifiers that significantly reduce the number of input pixels into an electronic network using an ultra-compact front-end $D^2$NN with a layer-to-layer distance of a few wavelengths, also reducing the complexity of the successive electronic network. Using a 5-layer phase-only $D^2$NN jointly-optimized with a single fully-connected electronic layer, we achieved a classification accuracy of 98.71% and 90.04% for the recognition of handwritten digits and fashion products, respectively. Moreover, the input to the electronic network was compressed by >7.8 times down to 10×10 pixels. Beyond creating low-power and high-frame rate machine learning platforms, $D^2$NN-based hybrid neural networks will find applications in smart optical imager and sensor design.

## I. Introduction

OPTICS in machine learning has been widely explored due to its unique advantages, encompassing power efficiency, speed and scalability[1]–[3]. Some of the earlier work include optical implementations of various neural network architectures[4]–[10], with a recent resurgence[11]–[22], following the availability of powerful new tools for applying deep neural networks[23], [24], which have redefined the state-of-the-art for a variety of machine learning tasks. In this line of work, we have recently introduced an optical machine learning framework, termed as Diffractive Deep Neural Network ($D^2$NN)[15], where deep learning and error back-propagation methods are used to design, using a computer, diffractive layers that collectively perform a desired task that the network is trained for. In this training phase of a $D^2$NN, the transmission and/or reflection coefficients of the individual pixels (i.e., neurons) of each layer are optimized such that as the light diffracts from the input plane toward the output plane, it computes the task at hand. Once this training phase in a computer is complete, these passive layers can be physically fabricated and stacked together to form an all-optical network that executes the trained function without the use of any power, except for the illumination light and the output detectors.

In our previous work, we experimentally demonstrated the success of $D^2$NN framework at THz part of the electromagnetic spectrum and used a standard 3D-printer to fabricate and assemble together the designed $D^2$NN layers[15]. In addition to demonstrating optical classifiers, we also demonstrated that the same $D^2$NN framework can be used to design an imaging system by 3D-engineering of optical components using deep learning[15]. In these earlier results, we used coherent illumination and encoded the input information in phase or amplitude channels of different $D^2$NN systems. Another important feature of $D^2$NNs is that the axial spacing between the diffractive layers is very small, e.g., less than 50 wavelengths ($\lambda$)[15], which makes the entire design highly compact and flat.

Our experimental demonstration of $D^2$NNs was based on linear materials, without including the equivalent of a nonlinear activation function within the optical network; however, as detailed in [15], optical nonlinearities can also be incorporated into a $D^2$NN using non-linear materials including e.g., crystals, polymers or semiconductors, to potentially improve its inference performance using nonlinear optical effects within diffractive layers. For such a nonlinear $D^2$NN design, resonant nonlinear structures (based on e.g., plasmonics or metamaterials) tuned to the illumination wavelength could be important to lower the required intensity levels. Even using linear optical materials to create a $D^2$NN, the optical network designed by deep learning shows "*depth*" advantage, i.e., a single diffractive layer does not possess the same degrees-of-freedom to achieve the same level of classification accuracy, power efficiency and signal contrast at the output plane that multiple diffractive layers can collectively achieve for a given task. It is true that, for a linear diffractive optical network, the entire wave propagation and diffraction phenomena that happen between the input and output planes can be squeezed into a *single* matrix operation; *however*, this arbitrary mathematical operation defined by multiple learnable diffractive layers cannot be performed in general by a single diffractive layer placed between the same input and output planes. That is why, multiple diffractive layers forming a D$^2$NN show the depth advantage, and statistically perform better compared to a single diffractive layer trained for the same classification task, and achieve improved accuracy as also discussed in the supplementary materials of [15].

Here, we present a detailed analysis of D$^2$NN framework, covering different parameters of its design space, also investigating the advantages of using multiple diffractive layers, and provide significant improvements to its inference performance by changing the loss function involved in the training phase, and reducing the effect of vanishing gradients in the error back-propagation step through its layers. To provide examples of its improved inference performance, using a 5-layer D$^2$NN design (Fig. 1), we optimized two different classifiers to recognize (1) hand-written digits, 0 through 9, using the MNIST (Mixed National Institute of Standards and Technology) image dataset[25], and (2) various fashion products, including t-shirts, trousers, pullovers, dresses, coats, sandals, shirts, sneakers, bags, and ankle boots (using the Fashion MNIST image dataset[26]). These 5-layer phase-only all-optical diffractive networks achieved a numerical blind testing accuracy of 97.18% and 89.13% for hand-written digit classification and fashion product classification, respectively. Using the same D$^2$NN design, this time with both the phase and the amplitude of each neuron's transmission as learnable parameters (which we refer to as *complex-valued* D$^2$NN design), we improved the inference performance to 97.81% and 89.32% for hand-written digit classification and fashion product classification, respectively. We also provide comparative analysis of D$^2$NN performance as a function of our design parameters, covering the impact of the number of layers, layer-to-layer connectivity and loss function used in the training phase on the overall classification accuracy, output signal contrast and power efficiency of D$^2$NN framework.

Furthermore, we report the integration of D$^2$NNs with electronic neural networks to create hybrid machine learning and computer vision systems. Such a hybrid system utilizes a D$^2$NN at its front-end, before the electronic neural network, and if it is jointly optimized (i.e., optical and electronic as a monolithic system design), it presents several important advantages. This D$^2$NN-based hybrid approach can all-optically *compress* the needed information by the electronic network using a D$^2$NN at its front-end, which can then significantly reduce the number of pixels (detectors) that needs to be digitized for an electronic neural network to act on. This would further improve the frame-rate of the entire system, also reducing the complexity of the electronic network and its power consumption. This D$^2$NN-based hybrid design concept can potentially create ubiquitous and low-power machine learning systems that can be realized using relatively simple and compact imagers, with e.g., a few tens to hundreds of pixels at the opto-electronic sensor plane, preceded by an ultra-compact all-optical diffractive network with a layer-to-layer distance of a few wavelengths, which presents important advantages compared to some other hybrid network configurations involving e.g., a 4-f configuration[16] to perform a convolution operation before an electronic neural network.

To better highlight these unique opportunities enabled by D$^2$NN-based hybrid network design, we conducted an analysis to reveal that a 5-layer phase-only (or *complex-valued*) D$^2$NN that is jointly-optimized with a single fully-connected layer, following the optical diffractive layers, achieves a blind classification accuracy of 98.71% (or *98.29%*) and 90.04% (or *89.96%*) for the recognition of hand-written digits and fashion products, respectively. In these results, the input image to the electronic network (created by diffraction through the jointly-optimized front-end D$^2$NN) was also compressed by more than 7.8 times, down to 10×10 pixels, which confirms that a D$^2$NN-based hybrid system can perform competitive classification performance even using a relatively simple and one-layer electronic network that uses significantly reduced number of input pixels.

In addition to potentially enabling ubiquitous, low-power and high-frame rate machine learning and computer vision platforms, these hybrid neural networks which utilize D$^2$NN-based all-optical processing at its front-end will find other applications in the design of compact and ultra-thin optical imaging and sensing systems by merging fabricated D$^2$NNs with opto-electronic sensor arrays. This will create intelligent systems benefiting from various CMOS/CCD imager chips and focal plane arrays at different parts of the electromagnetic spectrum, merging the benefits of all-optical computation with simple and low-power electronic neural networks that can work with lower dimensional data, all-optically generated at the output of a jointly-optimized D$^2$NN design.

## II. RESULTS AND DISCUSSION

*A. Mitigating vanishing gradients in optical neural network training*

In D$^2$NN framework, each neuron has a complex transmission coefficient, i.e., $t_i^l(x_i, y_i, z_i) = a_i^l(x_i, y_i, z_i) exp(j\phi_i^l(x_i, y_i, z_i))$, where $i$ and $l$ denote the neuron and diffractive layer number, respectively. In [15], $a_i^l$ and $\phi_i^l$ are represented during the network training as functions of two latent variables, $\alpha$ and $\beta$, defined in the following form:

$$a_i^l = sigmoid(\alpha_i^l), \quad (1a)$$

$$\phi_i^l = 2\pi \times sigmoid(\beta_i^l), \quad (1b)$$

where, $sigmoid(x) = \frac{e^x}{e^x+1}$, is a non-linear, differentiable function. In fact, the trainable parameters of a D$^2$NN are these latent variables, $\alpha_i^l$ and $\beta_i^l$, and eq. (1) defines how they are related to the physical parameters ($a_i^l$ and $\phi_i^l$) of a diffractive optical network. Note that in eq. (1), the sigmoid acts on an auxiliary variable rather than the information flowing through the network. Being a bounded analytical function, sigmoid confines the values of $a_i^l$ and $\phi_i^l$ inside the intervals $(0,1)$ and $(0,2\pi)$, respectively. On the other hand, it is known that sigmoid function has vanishing gradient problem[27] due to its relatively flat tails, and when it is used in the context depicted in eq. (1), it can prevent the network to utilize the available dynamic range considering both the amplitude and phase terms of each neuron. To mitigate these issues, in this work we replaced eq. (1) as follows:

$$a_i^l = \frac{ReLU(\alpha_i^l)}{max_{0<i\leq M}\{ReLU(\alpha_i^l)\}}, \quad (2a)$$

$$\phi_i^l = 2\pi \times \beta_i^l, \quad (2b)$$

where ReLU refers to Rectified Linear Unit, and $M$ is the number of neurons per layer. Based on eq. (2), the phase term of each neuron, $\phi_i^l$, becomes unbounded, but since the $exp(j\phi_i^l(x_i,y_i,z_i))$ term is periodic (and bounded) with respect to $\phi_i^l$, the error back-propagation algorithm is able to find a solution for the task in hand. The amplitude term, $a_i^l$, on the other hand, is kept within the interval $(0,1)$ by using an explicit normalization step shown in eq. (2).

To exemplify the impact of this change *alone* in the training of an all-optical D²NN design, for a 5-layer, phase-only (*complex-valued*) diffractive optical network with an axial distance of 40×λ between its layers, the classification accuracy for Fashion-MNIST dataset increased from reported 81.13% (*86.33%*) to 85.40% (*86.68%*) following the above discussed changes in the parameterized formulation of the neuron transmission values compared to earlier results in [15]. We will report further improvements in the inference performance of an all-optical D²NN after the introduction of the loss function related changes into the training phase, which is discussed next.

We should note that although the results of this paper follow the formulation in eq. (2), it is also possible to parameterize complex modulation terms over the real and imaginary parts as in [28] and a formulation based on the Wirtinger derivatives can be used for error backpropagation.

*B. Effect of the learning loss function on the performance of all-optical diffractive neural networks*

Earlier work on D²NNs[15] reports the use of mean squared error (MSE) loss. An alternative loss function that can be used for the design of a D²NN is the cross-entropy loss[29], [30] (see the Methods section). Since minimizing the cross-entropy loss is equivalent to minimizing the negative log-likelihood (or maximizing the likelihood) of an underlying probability distribution, it is in general more suitable for classification tasks. Note that, cross-entropy acts on probability measures, which take values in the interval $(0,1)$ and the signals coming from the detectors (one for each class) at the output layer of a D²NN are not necessarily in this range; therefore, in the training phase, a *softmax* layer is introduced to be able to use the cross-entropy loss. It is important to note that although *softmax* is used during the *training* process of a D²NN, once the diffractive design converges and is fixed, the class assignment at the output plane of a D²NN is still based *solely on the maximum optical signal detected at the output plane*, where there is one detector assigned for each class of the input data (see Figs. 1(a), 1(f)).

When we combine D²NN training related changes reported in the earlier sub-section on the parametrization of neuron modulation (eq. (2)), with the cross-entropy loss outlined above, a significant improvement in the classification performance of an all-optical diffractive neural network is achieved. For example, for the case of a 5-layer, phase-only D²NN with 40×λ axial distance between the layers, the classification accuracy for MNIST dataset increased from 91.75% to 97.18%, which further increased to 97.81% using complex-valued modulation, treating the phase and amplitude coefficients of each neuron as learnable parameters. The training convergence plots and the confusion matrices corresponding to these results are also reported in Figs. 2(a) and 2(c), for phase-only and complex-valued modulation cases, respectively. Similarly, for Fashion-MNIST dataset, we improved the blind testing classification accuracy of a 5-layer phase-only (*complex-valued*) D²NN from 81.13% (*86.33%*) to 89.13% (*89.32%*), showing a similar level of advancement as in the MNIST results. Figs. 3(a) and 3(c) also report the training convergence plots and the confusion matrices for these improved Fashion-MNIST inference results, for phase-only and complex-valued modulation cases, respectively. As a comparison point, a fully-electronic deep neural network such as ResNet-50[31] (with >25 Million learnable parameters) achieves 99.51% and 93.23% for MNIST and Fashion-MNIST datasets, respectively, which are superior to our 5-layer all-optical D²NN inference results (i.e., 97.81% and 89.32% for MNIST and Fashion-MNIST datasets, respectively), which in total used 0.8 million learnable parameters, covering the phase and amplitude values of the neurons at 5 successive diffractive layers.

All these results demonstrate that the D²NN framework using linear optical materials can already achieve a decent classification performance, also highlighting the importance of future research on the integration of optical nonlinearities into the layers of a D²NN, using e.g., plasmonics, metamaterials or other nonlinear optical materials (see the supplementary information of [15]), in order to come closer to the performance of state-of-the-art digital deep neural networks.

*C. Performance trade-offs in D²NN design*

Despite the significant increase observed in the blind testing accuracy of D²NNs, the use of softmax-cross-entropy (SCE) loss function in the context of all-optical networks also presents some trade-offs in terms of practical system parameters. MSE loss function operates based on pixel-by-pixel comparison of a user-designed output distribution with the output optical intensity pattern, after the input light interacts with the diffractive layers (see e.g., Figs. 1(d) and 1(i)). On the other hand, SCE loss function is much less restrictive for the spatial distribution or the uniformity of the output intensity at a given detector behind the diffractive layers (see e.g., Figs. 1(e) and 1(j)); therefore, it presents additional degrees-of-freedom and redundancy for the diffractive network to improve its inference accuracy for a given machine learning task, as reported in the earlier subsection.

This performance improvement with the use of SCE loss function in a diffractive neural network design comes at the expense of some compromises in terms of the expected diffracted power efficiency and signal contrast at the network output. To shed more light on this trade-off, we define the power efficiency of a $D^2NN$ as the percentage of the optical signal detected at the target label detector ($I_L$) corresponding to the correct data class with respect to the *total* optical signal at the output plane of the optical network ($E$). Fig. 4(b) and Fig. 4(e) show the power efficiency comparison as a function of the number of diffractive layers (corresponding to 1, 3 and 5-layer phase-only $D^2NN$ designs) for MNIST and Fashion-MNIST datasets, respectively. The power efficiency values in these graphs were computed as the ratio of the mean values of $I_L$ and $E$ for the test samples that were correctly classified by the corresponding $D^2NN$ designs (refer to Figs. 4(a) and 4(d) for the classification accuracy of each design). These results clearly indicate that increasing the number of diffractive layers has significant positive impact on the optical efficiency of a $D^2NN$, regardless of the loss function choice. The maximum efficiency that a 5-layer phase-only $D^2NN$ design based on the SCE loss function can achieve is 1.98% for MNIST and 0.56% for Fashion-MNIST datasets, which are significantly lower compared to the efficiency values that diffractive networks designed with MSE loss function can achieve, i.e., 25.07% for MNIST and 26.00% for Fashion-MNIST datasets (see Figs. 4(b) and 4(e)). Stated differently, MSE loss function based $D^2NNs$ are in general significantly more power efficient all-optical machine learning systems.

Next we analyzed the signal contrast of diffractive neural networks, which we defined as the difference between the optical signal captured by the target detector ($I_L$) corresponding to the correct data class and the maximum signal detected by the rest of the detectors (i.e., the strongest competitor ($I_{SC}$) detector for each test sample), normalized with respect to the total optical signal at the output plane ($E$). The results of our signal contrast analysis are reported in Fig. 4(c) and Fig. 4(f) for MNIST and Fashion-MNIST datasets, respectively, which reveal that $D^2NNs$ designed with an MSE loss function keep a strong margin between the target detector ($I_L$) and the strongest competitor detector (among the rest of the detectors) at the output plane of the all-optical network. The minimum mean signal contrast value observed for an MSE-based $D^2NN$ design was for a 1-Layer, phase-only diffractive design, showing a mean signal contrast of 2.58% and 1.37% for MNIST and Fashion-MNIST datasets, respectively. Changing the loss function to SCE lowers the overall signal contrast of diffractive neural networks as shown in Figs. 4(c) and 4(f).

Comparing the performances of MSE-based and SCE-based $D^2NN$ designs in terms of classification accuracy, power efficiency and signal contrast, as depicted in Fig. 4, we identify two opposite design strategies in diffractive all-optical neural networks. MSE, being a strict loss function acting in the physical space (e.g., Figs. 1(d) and 1(i)), promotes high signal contrast and power efficiency of the diffractive system, while SCE, being much less restrictive in its output light distribution (e.g., Figs. 1(e) and 1(j)), enjoys more degrees-of-freedom to improve its inference performance for getting better classification accuracy, at the cost of a reduced overall power efficiency and signal contrast at its output plane, which increases the systems' vulnerability for opto-electronic detection noise. In addition to the noise at the detectors, mechanical misalignment in both the axial and lateral directions might cause inference discrepancy between the final network model and its physical implementation. One way to mitigate this alignment issue is to follow the approach in Ref. [15] where the neuron size was chosen to be >3-4 times larger than the available fabrication resolution. Recently developed micro- and nano-fabrication techniques, such as laser lithography based on two-photon polymerization [32], emerge as promising candidates towards monolithic fabrication of complicated volumetric structures, which might help to minimize the alignment challenges in diffractive optical networks. Yet, another method of increasing the robustness against mechanical fabrication and related alignment errors is to model and include these error sources as part of the forward model during the numerical design phase, which might create diffractive models that are more tolerant of such errors.

*D. Advantages of multiple diffractive layers in $D^2NN$ framework*

As demonstrated in Fig. 4, multiple diffractive layers that collectively operate within a $D^2NN$ design present additional degrees-of-freedom compared to a single diffractive layer to achieve better classification accuracy, as well as improved diffraction efficiency and signal contrast at the output plane of the network; the latter two are especially important for experimental implementations of all-optical diffractive networks as they dictate the required illumination power levels as well as signal-to-noise ratio related error rates for all-optical classification tasks. Stated differently, $D^2NN$ framework, even when it is composed of linear optical materials, shows depth advantage because an increase in the number of diffractive layers (1) improves its statistical inference accuracy (see Figs. 4(a) and 4(d)), and (2) improves its overall power efficiency and the signal contrast at the correct output detector with respect to the detectors assigned to other classes (see Figs. 4(b), (c), (e), (f)). Therefore, for a given input illumination power and detector signal-to-noise ratio, the overall error rate of the all-optical network decreases as the number of diffractive layers increase. All these highlight the depth feature of a $D^2NN$.

This is not in contradiction with the fact that, for an all-optical $D^2NN$ that is made of linear optical materials, the entire diffraction phenomenon that happens between the input and output planes can be squeezed into a single matrix operation (in reality, every material exhibits some volumetric and surface nonlinearities, and what we mean here by a linear optical material is that these effects are negligible). In fact, such an arbitrary mathematical operation defined by multiple learnable diffractive layers cannot be performed in general by a single diffractive layer placed between the same input and output planes; additional optical components/layers would be needed to all-optically perform an arbitrary mathematical operation that multiple learnable diffractive layers can in general perform. Our D$^2$NN framework creates a unique opportunity to use deep learning principles to design multiple diffractive layers, within a very tight layer-to-layer spacing of less than 50×λ, that collectively function as an all-optical classifier, and this framework will further benefit from nonlinear optical materials[15] and resonant optical structures to further enhance its inference performance.

In summary, the "depth" is a feature/property of a neural network, which means the network gets in general better at its inference and generalization performance with more layers. The mathematical origins of the depth feature for standard electronic neural networks relate to nonlinear activation function of the neurons. But this is not the case for a diffractive optical network since it is a different type of a network, not following the same architecture or the same mathematical formalism of an electronic neural network.

*E. Connectivity in diffractive neural networks*

In a D$^2$NN design, the layer-to-layer connectivity of the optical network is controlled by several parameters: the axial distance between the layers ($\Delta_z$), the illumination wavelength (λ), the size of each fabricated neuron and the width of the diffractive layers. In our numerical simulations, we used a neuron size of approximately 0.53×λ. In addition, the height and width of each diffractive layer was set to include $200 \times 200 = 40K$ neurons per layer. In this arrangement, if the axial distance between the successive diffractive layers is set to be ~40×λ as in [15], then our D$^2$NN design becomes fully-connected. On the other hand, one can also design a much thinner and more compact diffractive network by reducing $\Delta_z$ at the cost of limiting the connectivity between the diffractive layers. To evaluate the impact of this reduction in network connectivity on the inference performance of a diffractive neural network, we tested the performance of our D$^2$NN framework using $\Delta_z = 4 \times \lambda$, i.e., 10-fold thinner compared to our earlier discussed diffractive networks. With this partial connectivity between the diffractive layers, the blind testing accuracy for a 5-layer, phase-only D$^2$NN decreased from 97.18% ($\Delta_z = 40 \times \lambda$) to 94.12% ($\Delta_z = 4 \times \lambda$) for MNIST dataset (see Figs. 2(a) and 2(b), respectively). However, when the optical neural network with $\Delta_z = 4 \times \lambda$ was relaxed from phase-only modulation constraint to full complex modulation, the classification accuracy increased to 96.01% (Fig. 2(d)), partially compensating for the lack of full-connectivity. Similarly, for Fashion-MNIST dataset, the same compact architecture with $\Delta_z = 4 \times \lambda$ provided accuracy values of 85.98% and 88.54% for phase-only and complex-valued modulation schemes, as shown in Figs. 3(b) and 3(d), respectively, demonstrating the vital role of phase and amplitude modulation capability for partially-connected, thinner and more compact optical networks (see the all-optical part of Table A2 in Appendix A).

*F. Integration of diffractive neural networks with electronic networks: Performance analysis of D$^2$NN-based hybrid machine learning systems*

Integration of passive diffractive neural networks with electronic neural networks (see e.g., Figs. 5(a) and 5(c)) creates some unique opportunities to achieve pervasive and low-power machine learning systems that can be realized using simple and compact imagers, composed of e.g., a few tens to hundreds of pixels per opto-electronic sensor frame. To investigate these opportunities, for both MNIST (Table I) and Fashion-MNIST (Table II) datasets, we combined our D$^2$NN framework (as an all-optical *front-end*, composed of 5 diffractive layers) with 5 different electronic neural networks considering various sensor resolution scenarios as depicted in Table III. For the electronic neural networks that we considered in this analysis, in terms of complexity and the number of trainable parameters, a single fully-connected (FC) digital layer and a custom designed 4-layer convolutional neural network (CNN) (we refer to it as 2C2F-1 due to the use of 2 convolutional layers with a single feature and subsequent 2 FC layers) represent the lower end of the spectrum (see Tables III-IV); on the other hand, LeNet[25], ResNet-50[31] and another 4-layer CNN[33] (we refer to it as 2C2F-64 pointing to the use of 2 convolutional layers, subsequent 2 FC layers and 64 high-level features at its second convolutional layer) represent some of the well-established and proven deep neural networks with more advanced architectures and considerably higher number of trainable parameters (see Table III). All these digital networks used in our analysis, were individually placed after both a fully-connected ($\Delta_z = 40 \times \lambda$) and a partially-connected ($\Delta_z = 4 \times \lambda$) D$^2$NN design and the entire hybrid system in each case was *jointly* optimized at the second stage of the hybrid system training procedure detailed in the Methods section (see Appendix A, Fig. A1).

Among the all-optical D$^2$NN-based classifiers presented in the previous sections, the fully-connected ($\Delta_z = 40 \times \lambda$) complex modulation D$^2$NN designs have the highest classification accuracy values, while the partially-connected ($\Delta_z = 4 \times \lambda$) designs with phase-only restricted modulation are at the bottom of the performance curve (see the *all-optical* parts of Tables I and II). Comparing the all-optical classification results based on a simple *max* operation at the output detector plane against the first rows of the "Hybrid Systems" sub-tables reported in Tables I and II, we can conclude that the addition of a single FC layer (using 10 detectors), jointly-optimized with the optical part, can make up for some of the limitations of the D$^2$NN optical front-end design such as partial connectivity or restrictions on the neuron modulation function.

The 2$^{nd}$, 3$^{rd}$ and 4$^{th}$ rows of the "Hybrid Systems" sub-tables reported in Tables I and II illustrate the classification performance of hybrid systems when the interface between the optical and electronic networks is a conventional focal plane array (such as a CCD or CMOS sensor array). The advantages of our D²NN framework become more apparent for these cases, compared against traditional systems that have a conventional imaging optics-based front-end (e.g., a standard camera interface) followed by a digital neural network for which the classification accuracies are also provided at the bottom of Tables I and II. From these comparisons reported in Tables I and II, we can deduce that having a jointly-trained optical and electronic network improves the inference performance of the overall system using low-end electronic neural networks as in the cases of a single FC network and 2C2F-1 network; also see Table III for a comparison of the digital neural networks employed in this work in terms of (1) the number of trainable parameters, (2) FLOPs, and (3) energy consumption. For example, when the 2C2F-1 network is used as the digital processing unit following a perfect imaging optics, the classification accuracies for MNIST (Fashion-MNIST) dataset are held as 89.73% (76.83%), 95.50% (81.76%) and 97.13% (87.11%) for 10×10, 25×25 and 50×50 detector arrays, respectively. However, when the same 2C2F-1 network architecture is enabled to jointly-evolve with e.g., the phase-only diffractive layers in a D²NN front-end during the training phase, blind testing accuracies for MNIST (Fashion-MNIST) dataset significantly improve to 98.12% (89.55%), 97.83% (89.87%) and 98.50% (89.42%) for 10×10, 25×25 and $50 \times 50$ detector arrays, respectively. The classification performance improvement of the jointly-optimized hybrid system (diffractive + electronic network) over a perfect imager-based simple all-electronic neural network (e.g., 2C2F-1) is especially significant for 10×10 detectors (i.e., ~8.4% and ~12.7% for MNIST and Fashion-MNIST datasets, respectively). Similar performance gains are also achieved when single FC network is jointly-optimized with D²NN instead of a perfect imaging optics/camera interface, preceding the all-electronic network as detailed in Tables I and II. In fact, for some cases the classification performance of D²NN-based hybrid systems, e.g. 5-layer, phase-only D²NN followed by a single FC layer using any of the 10×10, 25×25 and 50×50 detectors arrays, shows a classification performance on par with a perfect imaging system that is followed by a more powerful, and energy demanding LeNet architecture (see Table III).

Among the 3 different detector array arrangements that we investigated here, 10×10 detectors represent the case where the intensity on the opto-electronic sensor plane is severely undersampled. Therefore, the case of $10 \times 10$ detectors represents a substantial loss of information for the imaging-based scenario (note that the original size of the objects in both image datasets is $28 \times 28$). This effect is especially apparent in Table II, for Fashion-MNIST, which represents a more challenging dataset for object classification task, in comparison to MNIST. According to Table II, for a computer vision system with a perfect camera interface and imaging optics preceding the opto-electronic sensor array, the degradation of the classification performance due to spatial undersampling varies between 3% to 5% depending on the choice of the electronic network. *However*, jointly-trained hybrid systems involving trainable diffractive layers maintain their classification performance even with ~7.8 times reduced number of input pixels (i.e., 10×10 pixels compared to the raw data, 28×28 pixels). For example, the combination of a fully-connected (40×λ layer-to-layer distance) D²NN optical front-end with 5 phase-only (complex) diffractive layers followed by LeNet provides 90.24% (90.24%) classification accuracy for fashion products using a $10 \times 10$ detector array, which shows improvement compared to 87.44% accuracy that LeNet alone provides following a perfect imaging optics, camera interface. A similar trend is observed for all the jointly-optimized D²NN-based hybrid systems, providing 3-5% better classification accuracy compared to the performance of all-electronic neural networks following a perfect imager interface with 10×10 detectors. Considering the importance of compact, thin and low-power designs, such D²NN-based hybrid systems with significantly reduced number of opto-electronic pixels and an ultra-thin all-optical D²NN front-end with a layer-to-layer distance of a few wavelengths cast a highly sought design to extend the applications of jointly-trained opto-electronic machine learning systems to various fields, without sacrificing their performance.

On the other hand, for designs that involve higher pixel counts and more advanced electronic neural networks (with higher energy and memory demand), our results reveal that D²NN based hybrid systems perform worse compared to the inference performance of perfect imager-based computer vision systems. For example, based on Tables I and II one can infer that using ResNet as the electronic neural network of the hybrid system with 50x50 pixels, the discrepancy between the two approaches (D²NN vs. perfect imager based front-end choices) is ~0.5% and ~4% for MNIST and Fashion-MNIST datasets, respectively, in favor of the perfect imager front-end. We believe this inferior performance of the jointly-optimized D²NN-based hybrid system (when higher pixel counts and more advanced electronic networks are utilized) is related to sub-optimal convergence of the diffractive layers in the presence of a powerful electronic neural network that is by and large determining the overall loss of the jointly-optimized hybrid network during the training phase. In other words, considering the lack of non-linear activation functions within the D²NN layers, a powerful electronic neural network at the back-end hinders the evolution of the optical front-end during training phase due to its relatively superior approximation capability. Some of the recent efforts in the literature to provide a better understanding of the inner workings of convolutional neural networks[34],[35] might help us to devise more efficient learning schemes to overcome this "shadowing" behavior in order to improve the inference performance of our jointly-optimized D²NN-based hybrid systems. Extending the fundamental design principles and methods behind diffractive optical networks to operate under spatially and/or temporally incoherent illumination is another intriguing research direction stimulated by this work, as most computer vision systems of today rely on incoherent ambient light conditions. Finally, the flexibility of the D²NN framework paves the way for broadening our design space in the future to metasurfaces and metamaterials through essential modifications in the parameterization of the optical modulation functions [36], [37].

## III. METHODS

### A. Diffractive neural network architecture

In our diffractive neural network model, the input plane represents the plane of the input object or its data, which can also be generated by another optical imaging system or a lens, e.g., by projecting an image of the object data. Input objects were encoded in amplitude channel (MNIST) or phase channel (Fashion-MNIST) of the input plane and were illuminated with a uniform plane wave at a wavelength of $\lambda$ to match the conditions introduced in [15] for all-optical classification. In the hybrid system simulations presented in Tables I and II, on the other hand, the objects in both datasets were represented as amplitude objects at the input plane, providing a fair comparison between the two tables. A hybrid system performance comparison table for phase channel encoded Fashion-MNIST data is also provided in Table A2 (as part of Appendix A), providing a comparison to [15].

Optical fields at each plane of a diffractive network were sampled on a grid with a spacing of ~$0.53\lambda$ in both $x$ and $y$ directions. Between two diffractive layers, the free-space propagation was calculated using the angular spectrum method[15]. Each diffractive layer, with a neuron size of $0.53\lambda \times 0.53\lambda$, modulated the incident light in phase and/or amplitude, where the modulation value was a trainable parameter and the modulation method (phase-only or complex) was a pre-defined design parameter of the network. The number of layers and the axial distance from the input plane to the first diffractive layer, between the successive diffractive layers, and from the last diffractive layer to the detector plane were also pre-defined design parameters of each network. At the detector plane, the output field intensity was calculated.

### B. Forward propagation model

The physical model in our diffractive framework does not rely on small diffraction angles or the Fresnel approximation and is not restricted to far-field analysis (Fraunhofer diffraction) [38], [39]. Following the Rayleigh-Sommerfeld equation, a single neuron can be considered as the secondary source of wave $w_i^l(x, y, z)$, which is given by:

$$w_i^l(x, y, z) = \frac{z - z_i}{r^2}\left(\frac{1}{2\pi r} + \frac{1}{j\lambda}\right)\exp\left(\frac{j2\pi r}{\lambda}\right) \quad (3)$$

where $r = \sqrt{(x - x_i)^2 + (y - y_i)^2 + (z - z_i)^2}$ and $j = \sqrt{-1}$. Treating the input plane as the 0th layer, then for $l^{th}$ layer ($l \geq 1$), the output field can be modeled as:

$$u_i^l(x, y, z) = w_i^l(x, y, z) \cdot t_i^l(x_i, y_i, z_i) \cdot \sum_k u_k^{l-1}(x_i, y_i, z_i)$$
$$= w_i^l(x, y, z) \cdot |A| \cdot e^{j\Delta\theta}, \quad (4)$$

where $u_i^l(x, y, z)$ denotes the output of the $i^{th}$ neuron on $l^{th}$ layer located at $(x, y, z)$, the $t_i^l$ denotes the complex modulation, i.e., $t_i^l(x_i, y_i, z_i) = a_i^l(x_i, y_i, z_i)exp(j\phi_i^l(x_i, y_i, z_i))$. In eq. (4), $|A|$ is the relative amplitude of the secondary wave, and $\Delta\theta$ refers to the additional phase delay due to the input wave at each neuron, $\sum_k u_k^{l-1}(x_i, y_i, z_i)$, and the complex-valued neuron modulation function, $t_i^l(x_i, y_i, z_i)$.

### C. Training loss function

To perform classification by means of all-optical diffractive networks with minimal post-processing (i.e., using only a $max$ operation), we placed discrete detectors at the output plane. The number of detectors ($D$) is equal to the number of classes in the target dataset. The geometrical shape, location and size of these detectors ($6.4\lambda \times 6.4\lambda$) were determined before each training session. Having set the detectors at the output plane, the final loss value ($L$) of the diffractive neural network is defined through two different loss functions and their impact on D²NN based classifiers were explored (see the *Results* section). The first loss function was defined using the mean squared error (MSE) between the output plane intensity, $S^{l+1}$, and the target intensity distribution for the corresponding label, $G^{l+1}$, i.e.,

$$L = \frac{1}{K}\sum_i^K\left(S_i^{l+1} - G_i^{l+1}\right)^2, \quad (5)$$

where $K$ refers to the total number of sampling points representing the entire diffraction pattern at the output plane.

The second loss function used in combination with our all-optical D²NN framework is the cross-entropy. To use the cross-entropy loss function, an additional *softmax* layer is introduced and applied on the detected intensities (only during the training phase of a diffractive neural network design). Since *softmax* function is *not* scale invariant[40], the measured intensities by D detectors at the output plane are normalized such that they lie in the interval (0,10) for each sample. With $I_l$ denoting the total optical signal impinging onto the $l^{th}$ detector at the output plane, the normalized intensities, $I'_l$, can be found by,

$$I'_l = \frac{I_l}{\max\{I_l\}} \times 10. \quad (6)$$

In parallel, the cross-entropy loss function can be written as follows:

$$L = -\sum_l^D g_l \log(p_l), \quad (7)$$

where $p_l = \frac{e^{I'_l}}{\sum_l^D e^{I'_l}}$ and $g_l$ refer to the $l^{th}$ element in the output of the *softmax* layer, and the $l^{th}$ element of the ground truth label vector, respectively.

A key difference between the two loss functions is already apparent from eq. (5) and eq. (7). While the MSE loss function is acting on the entire diffraction signal at the output plane of the diffractive network, the *softmax-cross-entropy* is applied to the detected optical signal values ignoring the optical field distribution outside of the detectors (one detector is assigned per class). This approach based on *softmax-cross-entropy* loss brings additional degrees-of-freedom to the diffractive neural network training process, boosting the final classification performance as discussed in the *Results* section, at the cost of reduced diffraction efficiency and signal contrast at the output plane.

For both the imaging optics-based and hybrid (D$^2$NN + electronic) classification systems presented in Tables I and II, the loss functions were also based on *softmax-cross-entropy*.

*D. Diffractive network training*

All neural networks (optical and/or digital) were simulated using Python (v3.6.5) and TensorFlow (v1.10.0, Google Inc.) framework. All-optical, hybrid and electronic networks were trained for 50 epochs using a desktop computer with a GeForce GTX 1080 Ti Graphical Processing Unit, GPU and Intel(R) Core (TM) i9-7900X CPU @3.30GHz and 64GB of RAM, running Windows 10 operating system (Microsoft).

Two datasets were used in the training of the presented classifiers: MNIST and Fashion-MNIST. Both datasets have 70,000 objects/images, out of which we selected 55,000 and 5,000 as training and validation sets, respectively. Remaining 10,000 were reserved as the test set. During the training phase, after each epoch we tested the performance of the current model in hand on the 5K validation set and upon completion of the 50$^{th}$ epoch, the model with the best performance on 5K validation set was selected as the final design of the network models. All the numbers reported in this work are blind testing accuracy results held by applying these selected models on the 10K test sets.

The trainable parameters in a diffractive neural network are the modulation values of each layer, which were optimized using a back-propagation method by applying the adaptive moment estimation optimizer (Adam)[41] with a learning rate of 10$^{-3}$. We chose a diffractive layer size of 200×200 neurons per layer, which were initialized with $\pi$ for phase values and 1 for amplitude values. The training time was approximately 5 hours for a 5-layer D$^2$NN design with the hardware outlined above.

*E. D2NN-based hybrid network design and training*

To further explore the potentials of D$^2$NN framework, we co-trained diffractive network layers together with digital neural networks to form hybrid systems. In these systems, the detected intensity distributions at the output plane of the diffractive network were taken as the input for the digital neural network at the back-end of the system.

To begin with, keeping the optical architecture and the detector arrangement at the output plane of the diffractive network same as in the all-optical case, a single fully-connected layer was introduced as an additional component (replacing the simplest *max* operations in an all-optical network), which maps the optical signal values coming from *D* individual detectors into a vector of the same size (i.e., the number of classes in the dataset). Since there are 10 classes in both MNIST and Fashion-MNIST datasets, this simple fully-connected digital structure brings additional 110 trainable variables (i.e., 100 coefficients in the weight matrix and 10 bias terms) into our hybrid system.

We have also assessed hybrid configurations that pair D$^2$NNs with CNNs, a more popular architecture than fully-connected networks for object classification tasks. In such an arrangement, when the optical and electronic parts are directly cascaded and jointly-trained, the inference performance of the overall hybrid system was observed to stagnate at a local minimum (see Appendix A, Tables A1 and A2). As a possible solution to this issue, we divided the training of the hybrid systems into two stages as shown in Fig. A1. In the first stage, the detector array was placed right after the D$^2$NN optical front-end, which was followed by an additional, virtual optical layer, acting as an all-optical classifier (see Fig. A1(a)). We emphasize that this additional optical layer *is not* part of the hybrid system at the end; instead it will be replaced by a digital neural network in the second stage of our training process. The sole purpose of two-stage training arrangement used for hybrid systems is to find a better initial condition for the D$^2$NN that precedes the detector array, which is the interface between the fully optical and electronic networks.

In the second stage of our training process, the already trained 5-layer D$^2$NN optical front-end (preceding the detector array) was cascaded and jointly-trained with a digital neural network. It is important to note that the digital neural network in this configuration was trained from scratch. This type of procedure "resembles" transfer learning, where the additional layers (and data) are used to augment the capabilities of a trained model[42].

Using the above described training strategy, we studied the impact of different configurations, by increasing the number of detectors forming an opto-electronic detector array, with a size of 10×10, 25×25 and 50×50 pixels. Having different pixel sizes (see Table III), all the three configurations (10×10, 25×25 and 50×50 pixels) cover the central region of approximately 53.3$\lambda$×53.3$\lambda$ at the output plane of the D$^2$NN. Note that each detector configuration represents different levels of spatial undersampling applied at the output plane of a D$^2$NN, with 10×10 pixels corresponding to the most severe case. For each detector configuration, the first stage of the hybrid system training, shown in Fig. A1(a) as part of Appendix A, was carried out for 50 epochs providing the initial condition for 5-layer D$^2$NN design before the joint-optimization phase at the second stage. These different initial optical front-end designs along with their corresponding detector configurations were then combined and jointly-trained with various digital neural network architectures, simulating different hybrid systems (see Fig. A1(b) and Fig 5). At the interface of optical and electronic networks, we introduced a batch normalization layer applied on the detected intensity distributions at the sensor.

For the digital part, we focused on five different networks representing different levels complexity regarding (1) the number of trainable parameters, (2) the number of FLOPs in the forward model and (3) the energy consumption; see Table III. This comparative analysis depicted in Table III on energy consumption assumes that 1.5pJ is needed for each multiply-accumulate (MAC)[43] and based on this assumption, the 4$^{th}$ column of Table III reports the energy needed for each network configuration to classify an input image. The first one of these digital neural networks was selected as a single fully-connected (FC) network connecting every pixel of detector array with each one of the 10 output classes, providing as few as 1,000 trainable parameters (see Table III for details). We also used the 2C2F-1 network as a custom designed CNN with 2 convolutional and 2 FC layers with only a single filter/feature at each convolutional layer (see Table IV). As our 3rd network, we used LeNet[25] which requires a certain input size of 32×32 pixels, thus the detector array values were resized using bilinear interpolation before being fed into the electronic neural network. The fourth network architecture that we used in our comparative analysis (i.e., 2C2F-64), as described in [33], has 2 convolutional and 2 fully-connected layers similar to the second network, but with 32 and 64 features at the first and second convolutional layers, respectively, and has larger FC layers compared to the 2C2F-1 network. Our last network choice was ResNet-50[31] with 50 layers, which was only jointly-trained using the 50×50 pixel detector configuration, the output of which was resized using bilinear interpolation to 224×224 pixels before being fed into the network. The loss function of the $D^2NN$-based hybrid system was calculated by cross-entropy, evaluated at the output of the digital neural network.

As in $D^2NN$-based hybrid systems, the objects were assumed to be purely amplitude modulating functions for perfect imager-based classification systems presented in Tables I and II; moreover, the imaging optics or the camera system preceding the detector array is assumed to be diffraction limited which implies that the resolution of the captured intensity at the detector plane is directly limited by the pixel pitch of the detector array. The digital network architectures and training schemes were kept identical to $D^2NN$-based hybrid systems to provide a fair comparison. Also, worth noting, no data augmentation techniques have been used for any of the networks presented in this manuscript.

*F. Details of $D^2NN$-based hybrid system training procedure*

We introduced a two-stage training pipeline for $D^2NN$-based hybrid classifiers as mentioned in the previous sub-section. The main reason behind the development of this two-stage training procedure stems from the unbalanced nature of the $D^2NN$-based hybrid systems, especially if the electronic part of the hybrid system is a powerful deep convolutional neural network (CNN) such as ResNet. Being the more powerful of the two and the latter in the information processing order, deep CNNs adapt and converge faster than $D^2NN$-based optical front-ends. Therefore, directly cascading and jointly-training $D^2NNs$ with deep CNNs offer a suboptimal solution on the classification accuracy of the overall hybrid system. In this regard, Tables A1 and A2 (in Appendix A) illustrate examples of such a direct training approach. Specifically, Table A1 contains blind testing accuracy results for amplitude channel encoded handwritten digits when $D^2NN$-based optical front-end and electronic networks were directly cascaded and jointly-trained. Table A2, on the other hand, shows the testing accuracy results for fashion-products which are encoded in the phase channel at the input plane.

Figure A1 illustrates the two-step training procedure for $D^2NN$-based hybrid system training, which was used for the results reported in Tables I and II. In the first step, we introduce the detector array model that is going to be the interface between the optical and the electronic networks. An additional virtual diffractive layer is placed right after the detector plane (see Appendix A, Fig. A1(a)). We model the detector array as an intensity sensor (discarding the phase information). Implementing such a detector array model with an *average pooling* layer which has *strides* as large as its kernel size on both directions, the detected intensity, $I_A$, is held at the focal plane array. In our simulations, the size of $I_A$ was 10×10, 25×25 or 50×50, depending on the choice of the detector array used in our design. To further propagate this information through the virtual 1-Layer optical classifier (Fig. A1(a)), $I_A$ is interpolated using the *nearest neighbour* method back to the object size at the input plane. Denoting this interpolated intensity as $I_A^{'}$, the propagated field is given by $\sqrt{I_A^{'}}$ (see Fig. A1(a)). It is important to note that the phase information at the output plane of the $D^2NN$ preceding the detector array is entirely discarded, thus the virtual classifier decides solely based on the measured intensity (or underlying amplitude) as it would be the case for an electronic network.

After training this model for 50 epochs, the layers of the diffractive network preceding the detector array are taken as the ***initial*** condition for the optical part in the second stage of our training process (see Fig. A1(b)). Starting from the parameters of these diffractive layers, the second stage of our training simply involves the ***simultaneous*** training of a $D^2NN$-based optical part and an electronic network at the back-end of the detector array bridging two modalities as shown in Fig. A1(b). In this second part of the training, the detector array model is kept identical with the first part and the electronic neural network is trained from scratch with optical and electronic parts having equal learning rates ($10^{-3}$).

APPENDIX A

Appendix A includes Tables A1 and A2 as well as Figure A1.

# Tables

**All-Optical**

|  | $\Delta_z = 40\times\lambda$ | $\Delta_z = 4\times\lambda$ |
|---|---|---|
| Phase only | 97.18 | 94.12 |
| Complex | 97.81 | 96.01 |

**Hybrid Systems**

| # of detectors | Optical Modulation | Single FC Layer | | 2C2F-1 | | LeNet | | 2C2F-64 | | ResNet | |
|---|---|---|---|---|---|---|---|---|---|---|---|
| 10 | Phase only | 97.65 | 93.12 | N/A | N/A | N/A | N/A | N/A | N/A | N/A | N/A |
|  | Complex | 98.02 | 95.96 | N/A | N/A | N/A | N/A | N/A | N/A | N/A | N/A |
| 10×10 | Phase only | 98.71 | 98.21 | 98.12 | 97.62 | 98.42 | 98.25 | 98.55 | 98.23 | N/A | N/A |
|  | Complex | 98.29 | 98.20 | 98.35 | 97.60 | 98.59 | 98.25 | 98.56 | 98.31 | N/A | N/A |
| 25×25 | Phase only | 98.80 | 96.89 | 97.83 | 98.26 | 98.77 | 98.10 | 98.86 | 98.13 | N/A | N/A |
|  | Complex | 98.64 | 97.50 | 98.37 | 98.14 | 98.62 | 98.10 | 98.57 | 98.18 | N/A | N/A |
| 50×50 | Phase only | 98.82 | 98.07 | 98.50 | 97.88 | 98.65 | 97.93 | 98.92 | 98.35 | 98.97 | 98.09 |
|  | Complex | 98.81 | 97.99 | 98.17 | 98.22 | 98.56 | 98.06 | 98.63 | 98.32 | 98.54 | 98.11 |

**Imaging Optics Based Classification Systems**

| # of detectors | Single FC Layer | 2C2F-1 | LeNet | 2C2F-64 | ResNet |
|---|---|---|---|---|---|
| 10×10 | 91.50 | 89.73 | 98.36 | 98.18 | N/A |
| 25×25 | 92.91 | 95.50 | 98.83 | 98.99 | N/A |
| 50×50 | 92.44 | 97.13 | 98.95 | 99.04 | 99.53 |

**Table I.** Blind testing accuracies (reported in percentage) for all-optical (D²NN only), D²NN and perfect imager-based hybrid systems used in this work for MNIST dataset. In the D²NN-based hybrid networks reported here, 5 different digital neural networks spanning from a single fully-connected layer to ResNet-50 were co-trained with a D²NN design, placed before the electronic neural network. All the electronic neural networks used ReLU as the nonlinear activation function, and all the D²NN designs were based on spatially and temporally coherent illumination and linear optical materials, with 5 diffractive layers. For a discussion on methods to incorporate optical nonlinearities in a diffractive neural network, refer to [15]. Yellow and blue colors refer to $\Delta_z = 40\times\lambda$ and $\Delta_z = 4\times\lambda$, respectively.

### All-Optical

|  | $\Delta_z = 40\times\lambda$ | $\Delta_z = 4\times\lambda$ |
|---|---|---|
| Phase only | 88.57 | 85.69 |
| Complex | 88.94 | 88.29 |

### Hybrid Systems

| # of detectors | Optical Modulation | Digital Neural Networks | | | | | | | | |
|---|---|---|---|---|---|---|---|---|---|---|
| | | Single FC Layer | | 2C2F-1 | | LeNet | | 2C2F-64 | | ResNet | |
| 10 | Phase only | 88.88 | 87.76 | N/A | N/A | N/A | N/A | N/A | N/A | N/A | N/A |
| | Complex | 89.57 | 88.40 | N/A | N/A | N/A | N/A | N/A | N/A | N/A | N/A |
| 10×10 | Phase only | 90.04 | 88.84 | 89.55 | 88.83 | 90.24 | 89.19 | 90.08 | 89.76 | N/A | N/A |
| | Complex | 89.96 | 88.88 | 89.26 | 89.43 | 90.24 | 89.55 | 89.92 | 89.88 | N/A | N/A |
| 25×25 | Phase only | 90.08 | 88.75 | 89.87 | 89.02 | 89.96 | 89.20 | 89.84 | 89.66 | N/A | N/A |
| | Complex | 90.25 | 88.57 | 89.94 | 89.50 | 89.79 | 89.64 | 89.75 | 89.83 | N/A | N/A |
| 50×50 | Phase only | 90.22 | 89.43 | 89.42 | 89.72 | 89.71 | 89.24 | 89.66 | 90.30 | 89.20 | 89.43 |
| | Complex | 89.54 | 89.45 | 90.11 | 89.79 | 89.74 | 89.76 | 89.29 | 90.45 | 89.29 | 89.40 |

### Imaging Optics Based Classification Systems

| # of detectors | Single FC Layer | 2C2F-1 | LeNet | 2C2F-64 | ResNet |
|---|---|---|---|---|---|
| 10×10 | 81.20 | 76.83 | 87.44 | 88.11 | N/A |
| 25×25 | 84.47 | 81.76 | 90.19 | 91.6 | N/A |
| 50×50 | 84.49 | 87.11 | 90.33 | 91.9 | 93.46 |

**Table II.** Blind testing accuracies (reported in percentage) for all-optical (D²NN only), D²NN and perfect imager-based hybrid systems used in this work for Fashion-MNIST dataset. In the D²NN-based hybrid networks reported here, 5 different digital neural networks spanning from a single fully-connected layer to ResNet-50 were co-trained with a D²NN design, placed before the electronic neural network. All the electronic neural networks used ReLU as the nonlinear activation function, and all the D²NN designs were based on spatially and temporally coherent illumination and linear materials, with 5 diffractive layers. For a discussion on methods to incorporate optical nonlinearities in a diffractive neural network, refer to [15]. Yellow and blue colors refer to $\Delta_z = 40\times\lambda$ and $\Delta_z = 4\times\lambda$, respectively. For the results reported in the all-optical part of this table, Fashion-MNIST objects were encoded in the amplitude channel of the input plane. When they are encoded in the phase channel (as in [15]), blind testing accuracies for a 5-Layer, phase-only (complex) D²NN classifier become 89.13% (89.32%) with $\Delta_z = 40\times\lambda$ and 85.98% (88.54%) with $\Delta_z = 4\times\lambda$ as reported in Table A2, as part of Appendix A.

| Digital Neural Networks | Trainable Parameters | FLOPs | Energy Consumption (J/image) | Detector Configuration |
|---|---|---|---|---|
| Single FC Layer | 1000 | 2000 | $1.5\times10^{-9}$ | 10×10 |
| | 6250 | 12500 | $9.5\times10^{-9}$ | 25×25 |
| | 25000 | 50000 | $3.8\times10^{-8}$ | 50×50 |
| 2C2F-1 | 615 | 3102 | $2.4\times10^{-9}$ | 10×10 |
| | 825 | 9048 | $7.0\times10^{-9}$ | 25×25 |
| | 3345 | 43248 | $3.3\times10^{-8}$ | 50×50 |
| LeNet[25] | 60840 | $1\times10^{6}$ | $7.5\times10^{-7}$ | 10×10 |
| | | | | 25×25 |
| | | | | 50×50 |
| 2C2F-64[33] | $3.3\times10^{5}$ | $3.1\times10^{6}$ | $2.4\times10^{-6}$ | 10×10 |
| | $2.4\times10^{6}$ | $2.5\times10^{7}$ | $1.9\times10^{-5}$ | 25×25 |
| | $9.5\times10^{6}$ | $8.7\times10^{7}$ | $6.5\times10^{-5}$ | 50×50 |
| ResNet[31] | $25.5\times10^{6}$ | $4\times10^{9}$ | $3\times10^{-3}$ | 50×50 |

**Table III.** Comparison of *electronic neural networks* in terms of the number of trainable parameters, FLOPs and energy consumption; these are compared as they are part of the D²NN-based hybrid networks reported in this work. These 5 digital neural networks are using ReLU as the nonlinear activation function at each neuron. Energy consumption numbers, given in J/image, illustrates the energy needed by the corresponding neural network to classify a single image. It was assumed that 1.5pJ is consumed for each MAC.

| | Network architecture | | | | | | | |
|---|---|---|---|---|---|---|---|---|
| Layer Type | Conv layer 1 | | | Conv layer 2 | | | FC layer 1 | FC layer 2 |
| Activation | ReLU | | | ReLU | | | ReLU | Softmax |
| Detector configuration | kernel | Feature map | Stride | kernel | Feature map | Stride | Number of neurons | Number of neurons |
| 10×10 | 6×6 | 1 | 1 | 3×3 | 1 | 1 | 30 | 10 |
| 25×25 | | | 2 | | | 2 | | |
| 50×50 | | | 2 | | | 2 | | |

**Table IV.** Parameters of the custom designed network architecture which we refer to as 2C2F-1. Also see Table III for other details and comparison to other electronic neural networks used in this work.

## All-Optical

|  | $\Delta_z = 40 \times \lambda$ | $\Delta_z = 4 \times \lambda$ |
|---|---|---|
| Phase only | 97.18 | 94.12 |
| Complex | 97.81 | 96.01 |

## Hybrid Systems

| # of detectors | Optical Modulation | Digital Neural Networks ||||||||||
|---|---|---|---|---|---|---|---|---|---|---|---|
|  |  | Single FC Layer || 2C2F-1 || LeNet || 2C2F-64 || ResNet ||
| 10 | Phase only | 97.65 | 93.12 | N/A | N/A | N/A | N/A | N/A | N/A | N/A | N/A |
|  | Complex | 98.02 | 95.96 | N/A | N/A | N/A | N/A | N/A | N/A | N/A | N/A |
| 10×10 | Phase only | 98.23 | 98.17 | 98.22 | 97.98 | 98.31 | 98.53 | 98.93 | 98.32 | N/A | N/A |
|  | Complex | 98.07 | 98.51 | 97.87 | 98.16 | 98.42 | 98.56 | 99.00 | 98.42 | N/A | N/A |
| 25×25 | Phase only | 98.28 | 98.06 | 98.47 | 98.25 | 98.70 | 98.34 | 98.75 | 98.36 | N/A | N/A |
|  | Complex | 98.33 | 97.90 | 98.45 | 98.39 | 98.37 | 98.74 | 98.92 | 98.55 | N/A | N/A |
| 50×50 | Phase only | 98.22 | 98.36 | 98.57 | 98.45 | 98.42 | 98.49 | 98.88 | 98.46 | 98.38 | 99.02 |
|  | Complex | 98.35 | 98.19 | 98.32 | 98.49 | 98.71 | 98.45 | 99.07 | 98.13 | 98.43 | 99.16 |

## Imaging Optics Based Classification Systems

| # of detectors | Single FC Layer | 2C2F-1 | LeNet | 2C2F-64 | ResNet |
|---|---|---|---|---|---|
| 10×10 | 91.50 | 89.73 | 98.36 | 98.18 | N/A |
| 25×25 | 92.91 | 95.50 | 98.83 | 98.99 | N/A |
| 50×50 | 92.44 | 97.13 | 98.95 | 99.04 | 99.53 |

**Table A1**. Blind testing accuracies (reported in percentage) for all-optical (D²NN only), D²NN and perfect imager-based hybrid systems used in this work for MNIST dataset. The 2-stage hybrid system training discussed in the Methods section was ***not*** used here. Instead, D²NN and 5 different digital neural networks were jointly-trained at the same time from scratch. All the electronic neural networks used ReLU as the nonlinear activation function, and all the D²NN designs were based on spatially and temporally coherent illumination and linear materials, with 5 diffractive layers. Yellow and blue colors refer to $\Delta_z = 40 \times \lambda$ and $\Delta_z = 4 \times \lambda$, respectively.

### All-Optical

|  | $\Delta_z = 40 \times \lambda$ | $\Delta_z = 4 \times \lambda$ |
|---|---|---|
| Phase only | 89.13 | 85.98 |
| Complex | 89.32 | 88.54 |

### Hybrid Systems

| # of detectors | Optical Modulation | Digital Neural Networks ||||||||||
| | | Single FC Layer || 2C2F-1 || LeNet || 2C2F-64 || ResNet ||
|---|---|---|---|---|---|---|---|---|---|---|---|
| 10 | Phase only | 88.40 | 85.54 | N/A | N/A | N/A | N/A | N/A | N/A | N/A | N/A |
|  | Complex | 88.69 | 88.84 | N/A | N/A | N/A | N/A | N/A | N/A | N/A | N/A |
| 10×10 | Phase only | 88.67 | 89.90 | 89.05 | 88.95 | 88.66 | 89.56 | 89.12 | 88.79 | N/A | N/A |
|  | Complex | 88.82 | 89.60 | 89.02 | 88.93 | 89.25 | 89.68 | 89.46 | 89.51 | N/A | N/A |
| 25×25 | Phase only | 88.88 | 89.37 | 88.71 | 88.63 | 88.27 | 89.64 | 89.16 | 88.50 | N/A | N/A |
|  | Complex | 89.45 | 89.82 | 89.06 | 88.80 | 89.03 | 89.79 | 89.47 | 88.99 | N/A | N/A |
| 50×50 | Phase only | 88.98 | 89.55 | 88.23 | 88.39 | 88.28 | 89.20 | 88.99 | 88.79 | 88.22 | 88.94 |
|  | Complex | 89.34 | 89.84 | 88.94 | 88.55 | 87.81 | 89.27 | 89.32 | 88.99 | 88.59 | 88.98 |

### Imaging Optics Based Classification Systems

| # of detectors | Single FC Layer | 2C2F-1 | LeNet | 2C2F-64 | ResNet |
|---|---|---|---|---|---|
| 10×10 | 81.20 | 76.83 | 87.44 | 88.11 | N/A |
| 25×25 | 84.47 | 81.76 | 90.19 | 91.6 | N/A |
| 50×50 | 84.49 | 87.11 | 90.33 | 91.9 | 93.46 |

**Table A2.** Blind testing accuracies (reported in percentage) for all-optical (D²NN only), D²NN and perfect imager-based hybrid systems used in this work for Fashion-MNIST dataset. The 2-step hybrid system training discussed in the Methods was *not* used here. Instead, D²NN and 5 different digital neural networks were jointly-trained at the same time from scratch. In addition, the objects were encoded in the *phase* channel (0-2π) at the input plane, same as in [15]. All the electronic neural networks used ReLU as the nonlinear activation function, and all the D²NN designs were based on spatially and temporally coherent illumination and linear materials, with 5 diffractive layers. Yellow and blue colors refer to $\Delta_z = 40 \times \lambda$ and $\Delta_z = 4 \times \lambda$, respectively.

**List of Figures:**

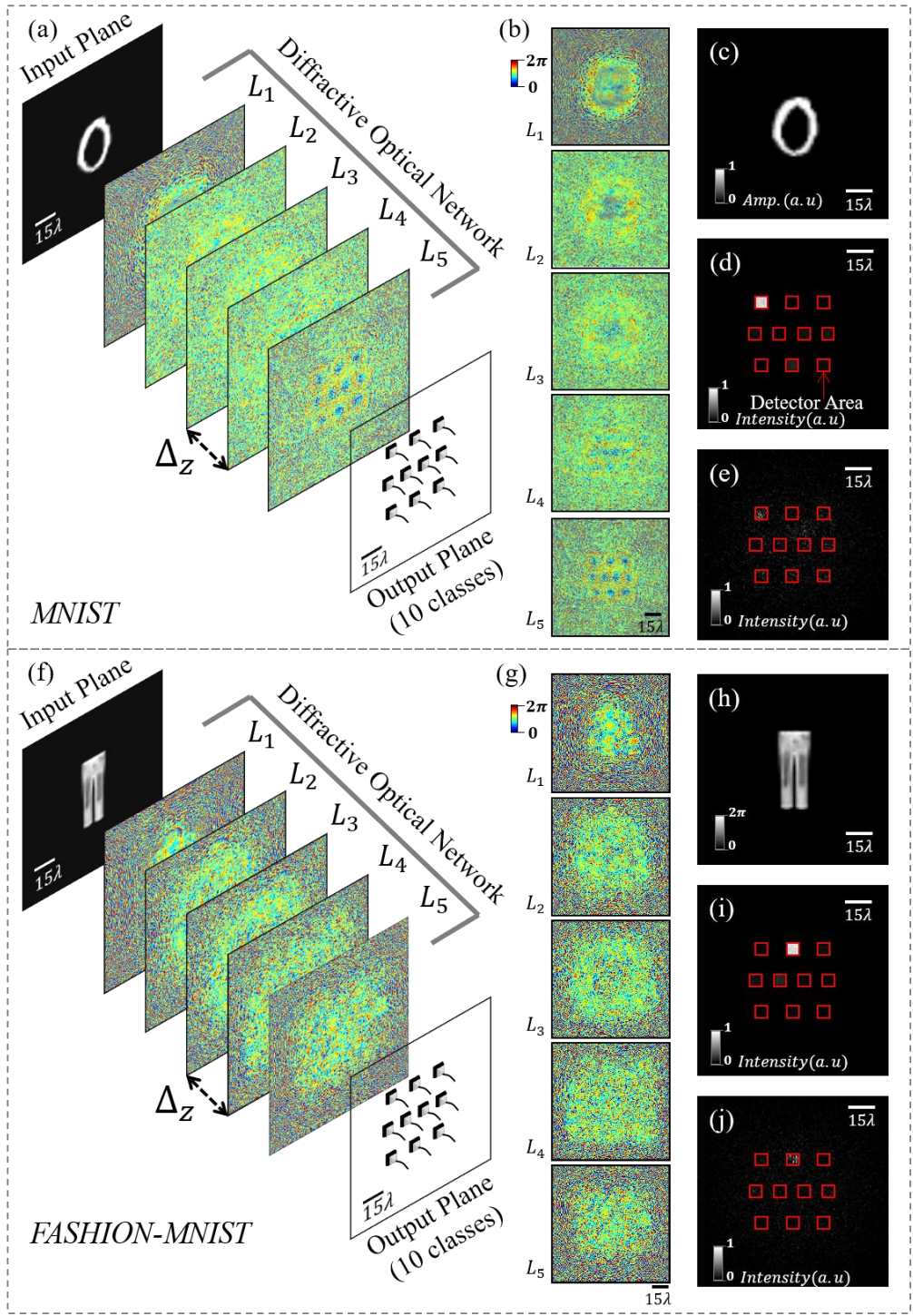

**Fig. 1.** All-optical $D^2NN$-based classifiers. These $D^2NN$ designs were based on spatially and temporally coherent illumination and linear optical materials/layers. (a) $D^2NN$ setup for the task of classification of handwritten digits (MNIST), where the input information is encoded in the *amplitude* channel of the input plane. (b) Final design of a 5-layer, phase-only classifier for handwritten digits. (c) Amplitude distribution at the input plane for a test sample (digit '0'). (d-e) Intensity patterns at the output plane for the input in (c); (d) is for MSE-based, and (e) is softmax-cross-entropy (SCE)-based designs. (f) $D^2NN$ setup for the task of classification of fashion products (Fashion-MNIST), where the input information is encoded in the *phase* channel of the input plane. (g) Same as (b), except for fashion product dataset. (h) Phase distribution at the input plane for a test sample. (i-j) Same as (d) and (e) for the input in (h). λ refers to the illumination source wavelength. Input plane represents the plane of the input object or its data, which can also be generated by another optical imaging system or a lens, projecting an image of the object data onto this plane.

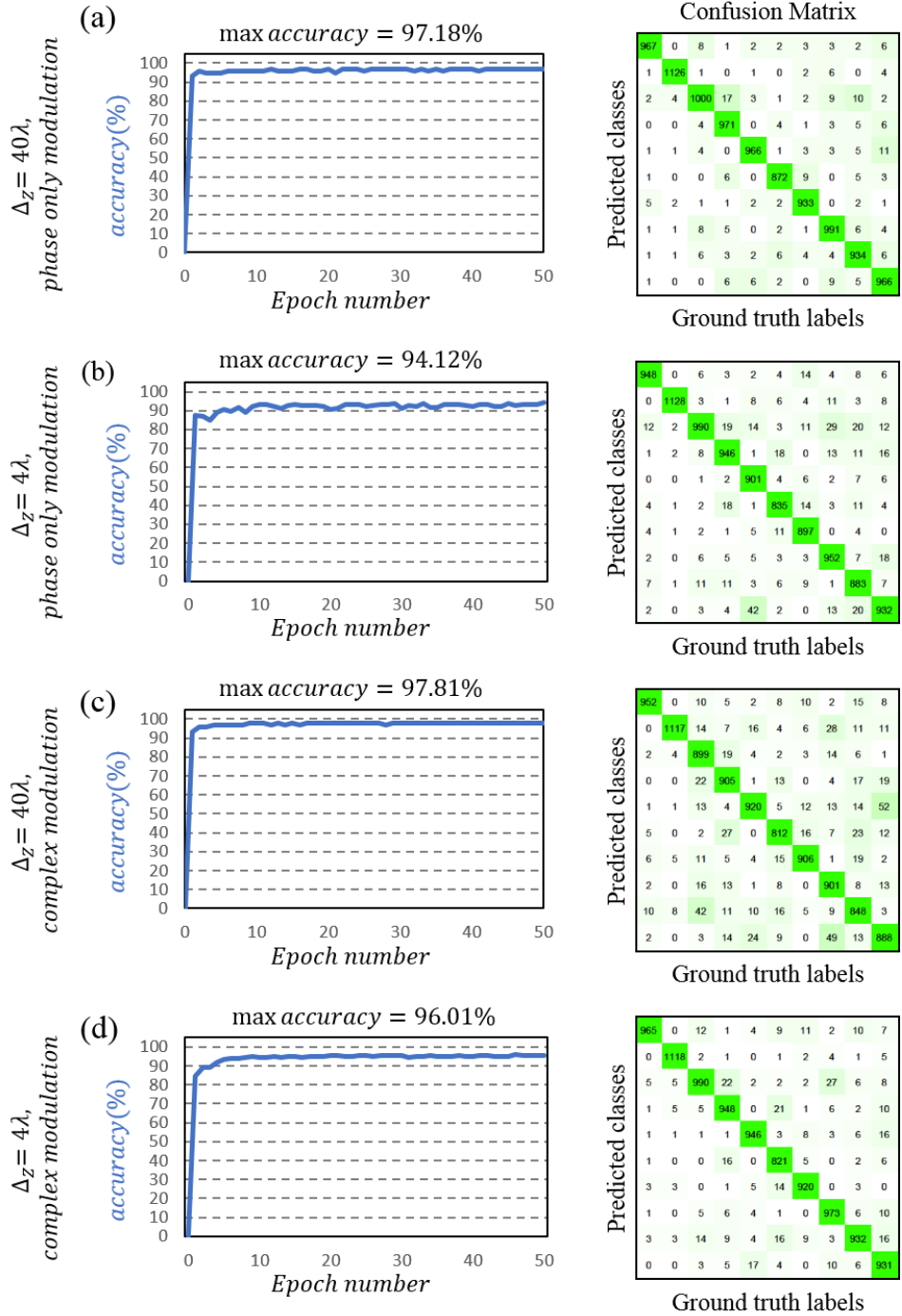

**Fig. 2.** Convergence plots and confusion matrices for all-optical $D^2NN$-based classification of handwritten digits (MNIST dataset). (a) Convergence curve and confusion matrix for a phase-only, fully-connected $D^2NN$ ($\Delta_z = 40\lambda$) design. (b) Convergence curve and confusion matrix for a phase-only, partially-connected $D^2NN$ ($\Delta_z = 4\lambda$) design. (c) and (d) are counterparts of (a) and (b), respectively, for complex-modulation $D^2NN$ designs, where both the amplitude and phase of each neuron are trainable parameters.

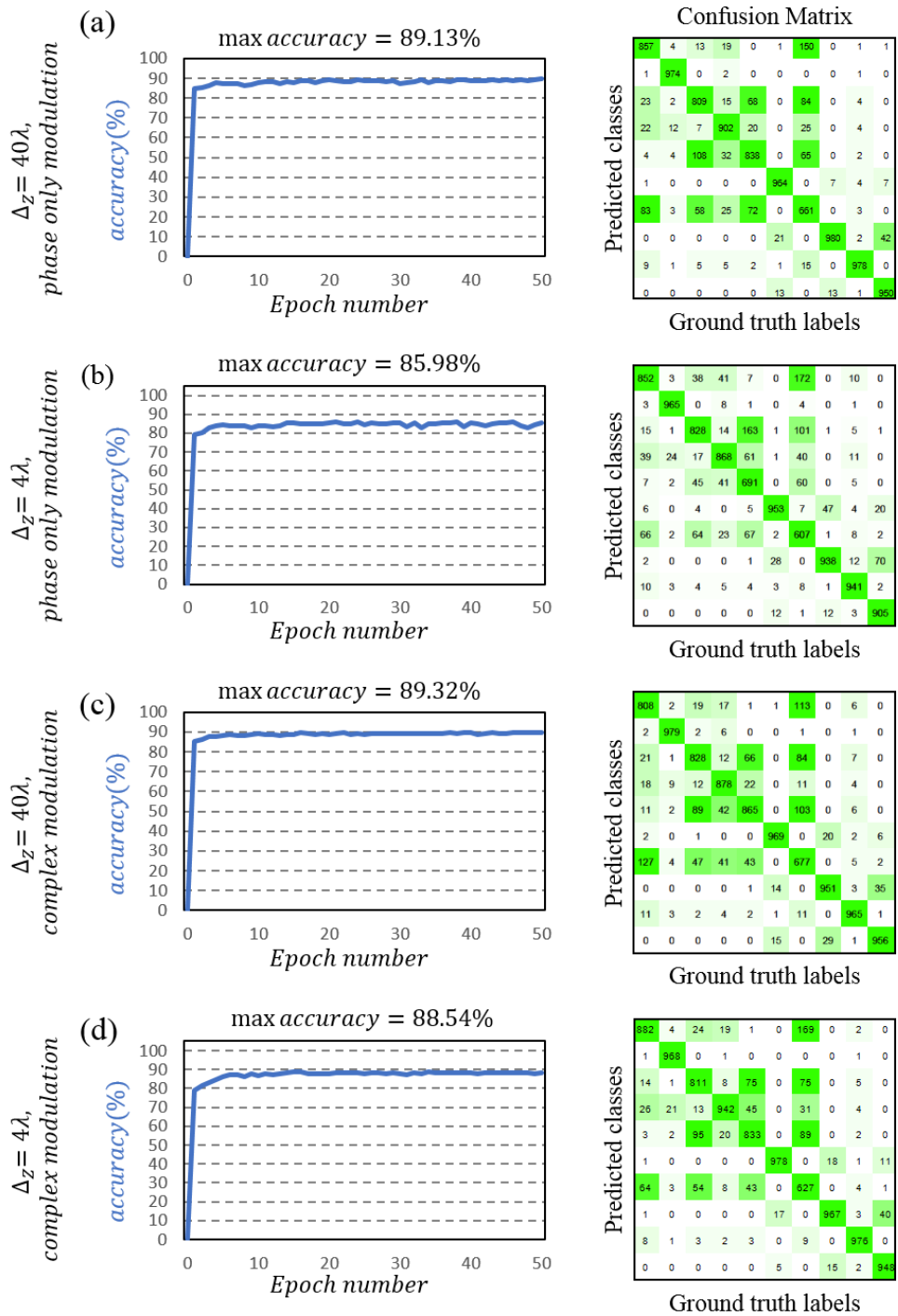

**Fig. 3.** Same as Fig. 2, except the results are for all-optical D$^2$NN-based classification of fashion products (Fashion-MNIST dataset) encoded in the phase channel of the input plane following [15].

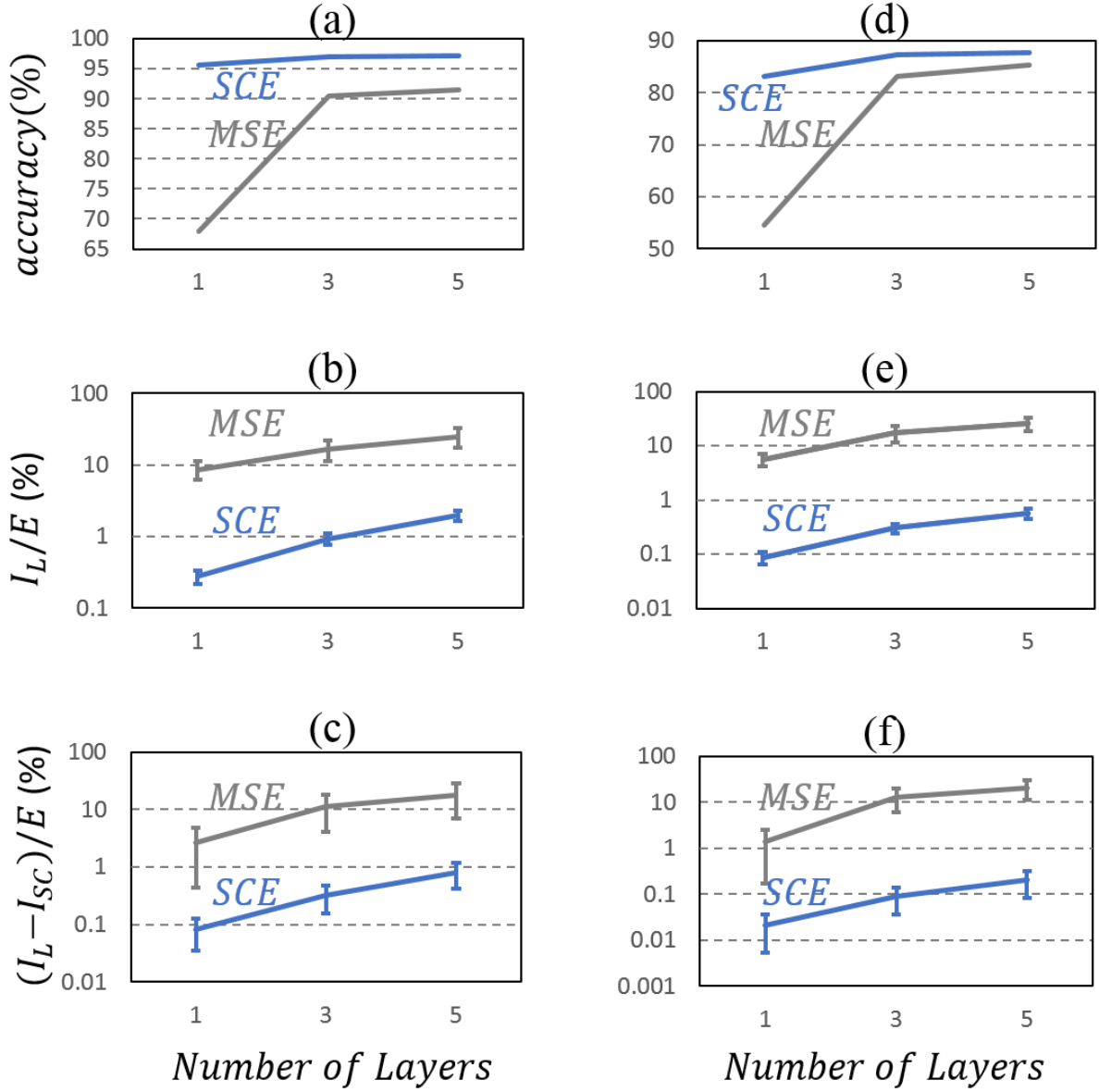

**Fig. 4.** Classification accuracy, power efficiency and signal contrast comparison of MSE and SCE loss function based all-optical phase-only D$^2$NN classifier designs with 1, 3 and 5-layers. (a) Blind testing accuracy, (b) power efficiency and (c) signal contrast analysis of the final design of fully-connected, phase-only all-optical classifiers trained for handwritten digits (MNIST). (d-f) are the same as (a-c), only the classified dataset is Fashion-MNIST instead.

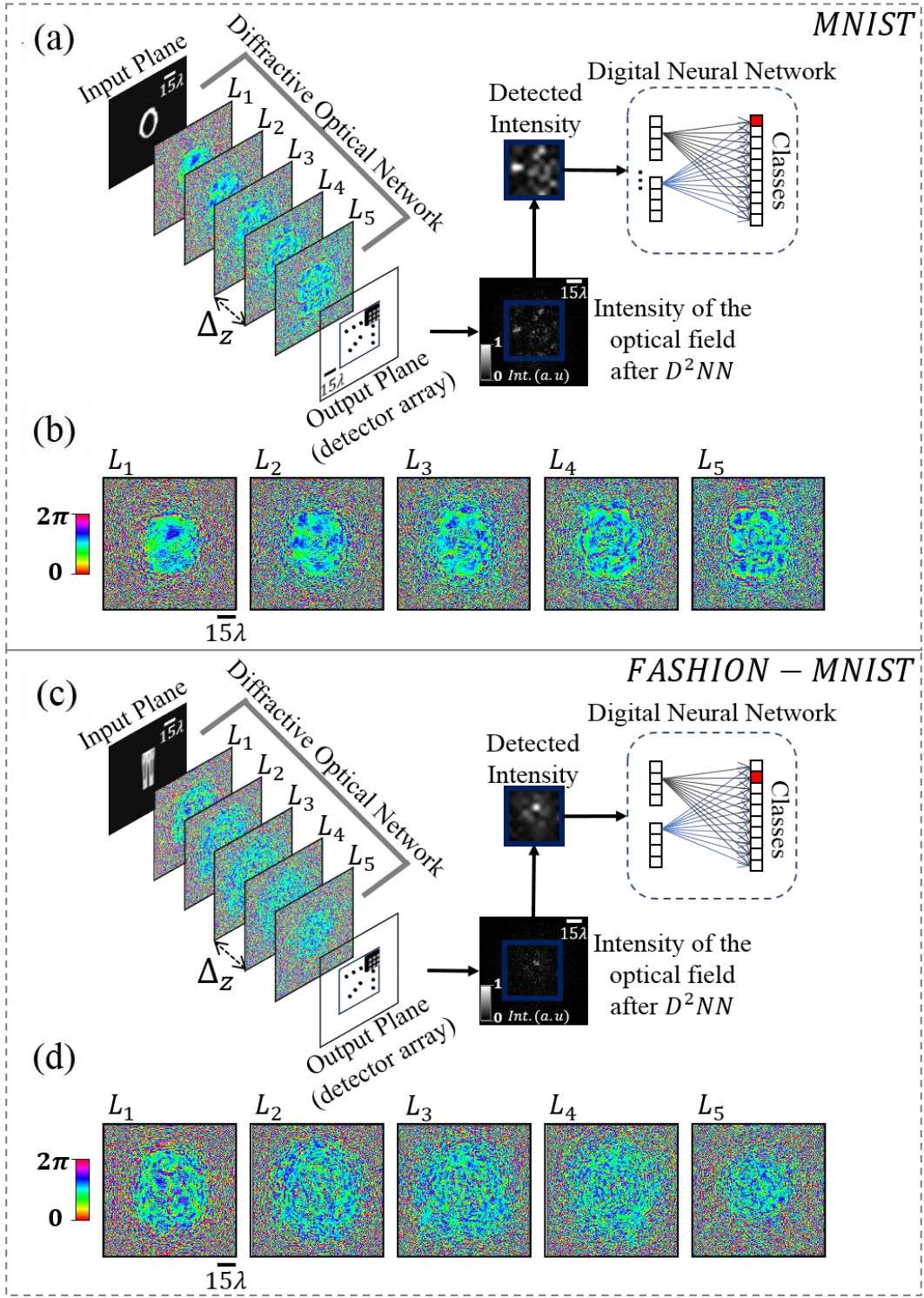

**Fig. 5.** D²NN-based hybrid neural networks. (a) The architecture of a hybrid (optical and electronic) classifier. (b) Final design of phase-only optical layers ($\Delta_z = 40 \times \lambda$) at the front-end of a hybrid handwritten digit classifier with a $10 \times 10$ opto-electronic detector array at the bridge/junction between the two modalities (optical vs. electronic). (c) and (d) are same as (a) and (b), except the latter are for Fashion-MNIST dataset. Input plane represents the plane of the input object or its data, which can also be generated by another optical imaging system or a lens, projecting an image of the object data onto this plane.

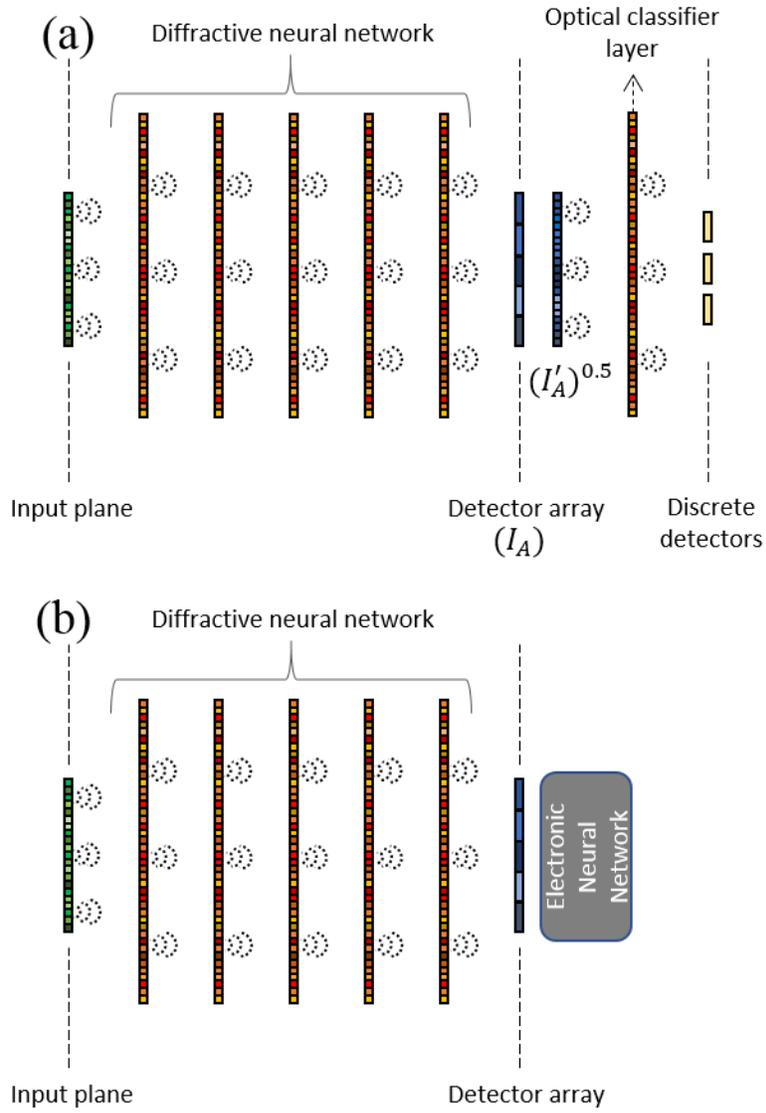

**Fig. A1.** Hybrid system training procedure. (a) The first stage of the hybrid system training. (b) The second stage of the hybrid system training starts with the already trained diffractive layers (first 5 layers) from part (a) and an electronic neural network, replacing the operations after intensity detection at the sensor. Note that the spherical waves between the consequent layers in (a) and (b) illustrate free space wave propagation.